\documentclass[conference]{IEEEtran}
\IEEEoverridecommandlockouts
\usepackage{cite}
\usepackage{amsmath,amssymb,amsfonts}
\usepackage{algorithmic}
\usepackage{graphicx}
\usepackage{textcomp}
\usepackage{xcolor}
\def\BibTeX{{\rm B\kern-.05em{\sc i\kern-.025em b}\kern-.08em
    T\kern-.1667em\lower.7ex\hbox{E}\kern-.125emX}}
\begin{document}

\title{GazeProphet: Software-Only Gaze Prediction for VR Foveated Rendering}

\author{
\IEEEauthorblockN{Farhaan Ebadulla\IEEEauthorrefmark{1}, 
Chiraag Mudlapur\IEEEauthorrefmark{1}, 
Gaurav BV\IEEEauthorrefmark{1}}
\IEEEauthorblockA{\IEEEauthorrefmark{1}Dept. of Computer Science and Engineering\\
PES University\\
Bengaluru, Karnataka, India\\
\{pes2202101091, pes2202101137, pes2202100517\}@pesu.pes.edu}
}

\maketitle

\begin{abstract}
Foveated rendering significantly reduces computational demands in virtual reality applications by concentrating rendering quality where users focus their gaze. Current approaches require expensive hardware-based eye tracking systems, limiting widespread adoption due to cost, calibration complexity, and hardware compatibility constraints. This paper presents GazeProphet, a software-only approach for predicting gaze locations in VR environments without requiring dedicated eye tracking hardware. The approach combines a Spherical Vision Transformer for processing 360-degree VR scenes with an LSTM-based temporal encoder that captures gaze sequence patterns. A multi-modal fusion network integrates spatial scene features with temporal gaze dynamics to predict future gaze locations. Experimental evaluation on a comprehensive VR dataset demonstrates that GazeProphet achieves a median angular error of 3.83 degrees, outperforming traditional saliency-based baselines by 24\%. The approach maintains consistent performance across different spatial regions and scene types, enabling practical deployment in VR systems without additional hardware requirements. Statistical analysis confirms the significance of improvements across all evaluation metrics. These results show that software-only gaze prediction can work for VR foveated rendering, making this performance boost more accessible to different VR platforms and apps.
\end{abstract}


\section{Introduction}

Virtual reality applications require substantial computational resources to render high-quality immersive experiences \cite{nvidia_vr_2024}. Modern VR systems often struggle to maintain acceptable frame rates while delivering detailed visual content \cite{oculus_rendering_2023}. This challenge becomes more pronounced as VR displays increase in resolution and field of view.

Foveated rendering offers a promising solution to this computational bottleneck \cite{guenter_foveated_2012, patney_towards_2016}. The technique reduces rendering quality in peripheral vision areas while maintaining high detail where users focus their gaze. Human visual perception naturally exhibits reduced acuity outside the central foveal region \cite{rosenholtz_capabilities_2016}. Foveated rendering exploits this characteristic to achieve significant performance improvements without noticeable quality degradation.

Current foveated rendering implementations rely on dedicated eye tracking hardware \cite{tobii_eye_2023, smieyetracking_2024}. These systems use infrared cameras and specialized sensors to monitor eye movements in real time. The hardware adds substantial cost to VR headsets. Installation requires precise calibration procedures. The additional components increase power consumption and system complexity.

Hardware-based eye tracking limits foveated rendering adoption across the VR ecosystem. Premium headsets like Meta Quest Pro and Apple Vision Pro include eye tracking capabilities \cite{meta_quest_pro_2022, apple_vision_pro_2023}. However, the majority of VR users own devices without this hardware. Popular headsets such as Meta Quest 2 and Quest 3 serve millions of users but lack eye tracking functionality. These users cannot benefit from foveated rendering performance optimizations.

Software-only gaze prediction presents an alternative approach \cite{krafka_eye_2016, cheng_gaze_2020}. The technique estimates gaze locations using scene content and temporal patterns without requiring additional hardware. Machine learning models can learn from user behavior patterns to predict where attention focuses during VR experiences \cite{dosovitskiy_attention_2021}. Recent advances in computer vision and deep learning make such predictions increasingly feasible.

This paper presents GazeProphet, a software-only approach for VR gaze prediction. The system combines spatial scene analysis with temporal gaze sequence modeling. A Spherical Vision Transformer processes 360-degree VR environments to extract spatial features. An LSTM-based encoder captures temporal patterns in gaze sequences. A multi-modal fusion network integrates these components to predict future gaze locations with confidence estimates.

The approach addresses several key challenges in VR gaze prediction. Spherical image geometry requires specialized processing techniques different from traditional computer vision approaches. Temporal dependencies in gaze patterns provide valuable predictive information. Confidence estimation enables adaptive foveated rendering based on prediction reliability.

Experimental evaluation demonstrates the effectiveness of this approach. Testing on comprehensive VR datasets shows median angular error of 3.83 degrees. The system outperforms traditional saliency-based baselines by 24 percent. Performance remains consistent across different spatial regions and scene types. Statistical analysis confirms the significance of improvements across evaluation metrics.

The contributions of this work include: (1) a novel software-only approach for VR gaze prediction that eliminates hardware requirements, (2) integration of spherical vision transformers with temporal sequence modeling for VR environments, and (3) comprehensive experimental validation showing practical performance levels for foveated rendering applications.

These results establish the feasibility of software-only gaze prediction for VR systems. The approach enables foveated rendering on existing hardware without additional costs. This democratizes access to performance optimization techniques across diverse VR platforms and applications.

\section{Related Work}

\subsection{Eye Tracking in VR Systems}

Eye tracking technology in virtual reality has evolved significantly over the past decade. Commercial systems like Tobii Pro VR Integration \cite{tobii_pro_vr_2023} and SMI Eye Tracking HMD \cite{smi_hmd_2022} provide high-precision gaze tracking for research applications. These systems achieve sub-degree accuracy but require specialized hardware integration.

Consumer VR headsets have begun incorporating eye tracking capabilities. The HTC Vive Pro Eye \cite{htc_vive_pro_eye_2019} introduced eye tracking to mainstream VR gaming. Meta Quest Pro \cite{meta_quest_pro_2022} includes advanced eye tracking sensors for social applications. Apple Vision Pro \cite{apple_vision_pro_2023} employs sophisticated eye tracking for user interface control. PlayStation VR2 \cite{sony_psvr2_2023} integrates eye tracking with foveated rendering support.

Despite these advances, eye tracking remains expensive and complex. Hardware costs add hundreds of dollars to headset prices. Calibration procedures require user training and frequent recalibration. Individual differences in eye anatomy affect tracking accuracy \cite{holmqvist_eye_2011}. Environmental factors like lighting conditions impact system performance \cite{clay_eye_2019}.

\subsection{Foveated Rendering Approaches}

Foveated rendering techniques aim to reduce computational load while maintaining visual quality. Early implementations used fixed foveated rendering \cite{murphy_application_2009}. These systems apply reduced quality to predefined peripheral regions without tracking actual gaze locations.

Dynamic foveated rendering adapts quality based on real-time gaze tracking \cite{guenter_foveated_2012, patney_towards_2016}. Research demonstrates significant performance improvements with minimal perceptual impact \cite{weier_foveated_2017}. Studies show 50-70\% rendering cost reductions while maintaining visual fidelity \cite{nvidia_vrs_2019}.

Recent work explores perceptually-guided foveated rendering \cite{steinicke_human_2021}. These approaches incorporate human visual system models to optimize quality allocation. Temporal aspects of vision receive attention through motion-aware foveated rendering \cite{albert_dynamic_2017}. Multi-resolution approaches balance quality and performance across different viewing conditions \cite{kaplanyan_deepfovea_2019}.

\subsection{Gaze Prediction and Saliency Models}

Visual attention prediction has extensive research history in computer vision. Traditional saliency models use hand-crafted features to identify attention-grabbing regions \cite{itti_model_1998, harel_graph_2007}. These approaches focus on low-level visual features like color, intensity, and orientation.

Deep learning transformed saliency prediction capabilities. Convolutional neural networks demonstrate superior performance over traditional methods \cite{kruthiventi_deepfix_2017, cornia_predicting_2018}. Attention mechanisms enhance model interpretability and performance \cite{wang_revisiting_2018}.

Video saliency prediction incorporates temporal information for dynamic scenes \cite{mathe_actions_2016, gorji_end_2019}. These models capture motion patterns and temporal dependencies in gaze behavior. Recurrent neural networks and LSTM architectures prove effective for sequence modeling \cite{linardos_simple_2021}.

Gaze prediction in 360-degree content presents unique challenges. Spherical image geometry requires specialized processing techniques \cite{xu_gaze_2018, chao_salnet360_2018}. VR environments require temporal modeling of gaze sequences for effective prediction \cite{david_dataset_2018}. Recent work addresses these challenges through spherical convolutions and attention mechanisms \cite{ling_saltivrnet_2022}.

\subsection{Vision Transformers and Attention Mechanisms}

Vision Transformers revolutionized computer vision with self-attention mechanisms \cite{dosovitskiy_attention_2021}. These models demonstrate strong performance across various visual tasks without convolutional layers. Patch-based processing enables flexible input handling and global context modeling.

Applications of Vision Transformers expand beyond traditional image classification. Object detection \cite{carion_end_2020}, semantic segmentation \cite{zheng_rethinking_2021}, and video understanding \cite{arnab_vivit_2021} benefit from transformer architectures. Recent work explores specialized vision transformers for spherical images \cite{cohen_spherical_2018, jiang_spherephd_2019}.

Attention mechanisms enhance model interpretability and performance in gaze prediction tasks. Multi-head attention captures different aspects of visual information \cite{vaswani_attention_2017}. Cross-modal attention enables integration of different information sources \cite{lu_12_2019}. These mechanisms prove particularly valuable for combining spatial and temporal information in sequence prediction tasks.

\subsection{Temporal Sequence Modeling}

Long Short-Term Memory networks excel at capturing temporal dependencies in sequential data \cite{hochreiter_long_1997}. LSTM architectures address vanishing gradient problems in recurrent neural networks. Applications in computer vision include video analysis \cite{donahue_long_2015} and sequential prediction tasks \cite{xingjian_convolutional_2015}.

Gaze sequence modeling benefits from temporal architectures. Previous gaze locations provide strong predictive signals for future attention \cite{zelinsky_eye_2013}. Temporal patterns vary across individuals and tasks, requiring adaptive modeling approaches \cite{henderson_high_2003}. Recent work combines convolutional and recurrent architectures for improved spatiotemporal modeling \cite{huang_salicon_2015}.

Multi-modal fusion techniques integrate different information sources for enhanced prediction. Early fusion combines features at input level, while late fusion merges predictions from separate models \cite{baltrusaitis_multimodal_2019}. Attention-based fusion enables adaptive weighting of different modalities based on context \cite{zadeh_tensor_2017}. These approaches prove effective for combining spatial scene content with temporal gaze patterns.

\begin{figure*}[htbp]
    \centering
    \includegraphics[width=0.7\textwidth]{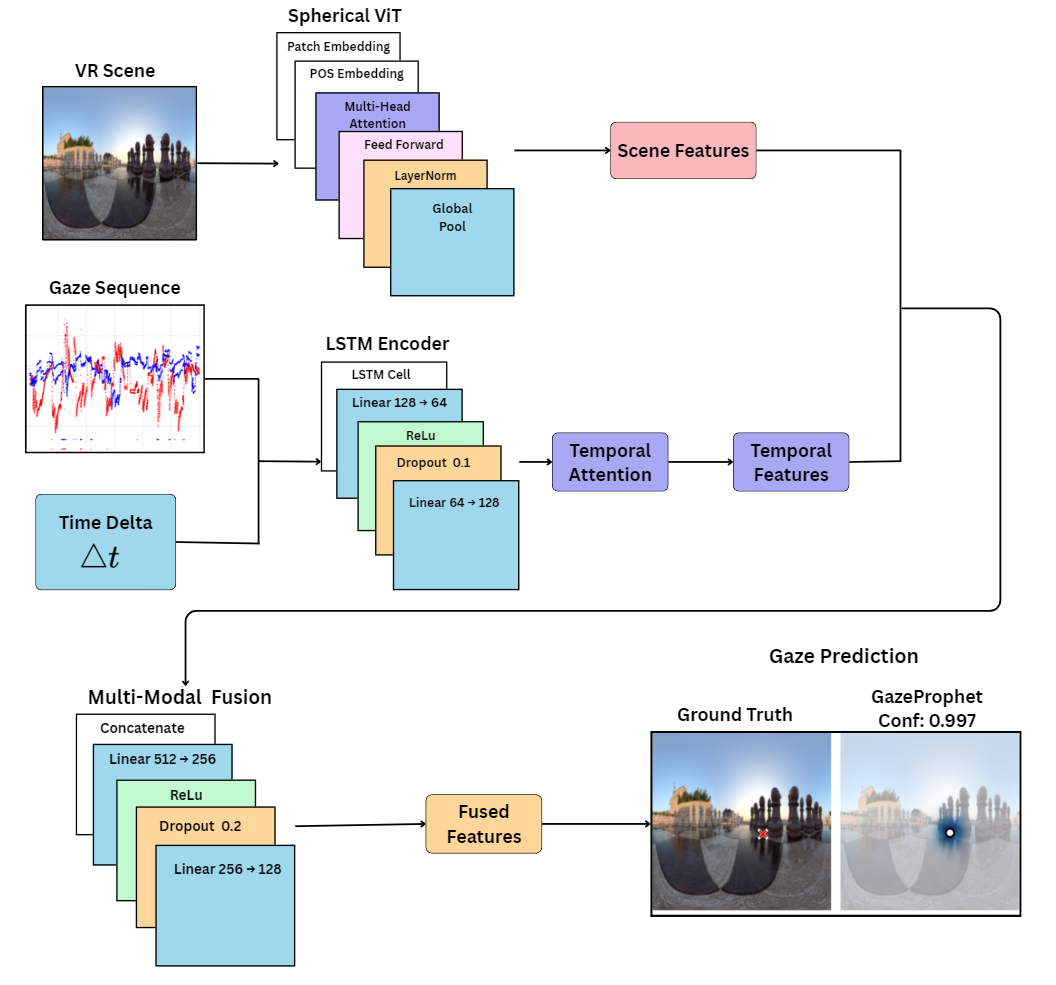}
    \caption{GazeProphet Architecture: Multi-modal approach combining Spherical Vision Transformer for 360° scene processing, LSTM Temporal Encoder for gaze sequence modeling, and Multi-Modal Fusion Network for final prediction generation.}
    \label{fig:architecture}
\end{figure*}

\section{Approach}

The GazeProphet approach combines spatial scene analysis with temporal gaze sequence modeling to predict future gaze locations in VR environments. The system processes 360-degree equirectangular images using a Spherical Vision Transformer while simultaneously analyzing gaze history through an LSTM-based temporal encoder. A multi-modal fusion network integrates these components and produces gaze predictions with confidence estimates.

\subsection{Architecture Overview}

Figure~\ref{fig:architecture} shows the complete system consisting of three main components working in parallel. The Spherical Vision Transformer extracts spatial features from VR scene images. The LSTM Temporal Encoder processes sequences of previous gaze points with timing information. The Multi-Modal Fusion Network combines spatial and temporal features to generate final predictions.

Input data includes a 360-degree VR scene image at 256×512 resolution and a sequence of 10 previous gaze points. Each gaze point contains $x,y$ coordinates in normalized $[0,1]$ range and a confidence value. Time delta information between consecutive gaze points provides temporal context.

The system outputs a 2D gaze prediction $(x,y$ coordinates$)$ and an associated confidence score for potential use in adaptive rendering strategies.

\subsection{Spherical Vision Transformer}

VR scenes require specialized processing due to equirectangular projection distortion. Standard Vision Transformers assume planar geometry and fail to handle spherical coordinate systems effectively. The Spherical Vision Transformer addresses these limitations through modified patch embedding and positional encoding.

\subsubsection{Patch Embedding Process}

The system divides 256×512 input images into 16×16 pixel patches. This creates a grid of 16×32 patches, totaling 512 patch tokens. Each patch undergoes linear projection to 384-dimensional feature vectors. The patch size balances computational efficiency with spatial resolution requirements.

Spherical coordinates require different treatment than standard Cartesian patches. The approach applies patch-wise normalization to account for varying pixel density across equirectangular images. Patches near image poles represent smaller spherical areas than equatorial patches. Normalization weights compensate for this geometric distortion.

\subsubsection{Positional Encoding Implementation}

Standard sinusoidal positional encoding fails for spherical coordinates. The system implements spherical harmonic-based positional encoding that respects the underlying spherical geometry. Each patch position $(i,j)$ maps to spherical coordinates $(\theta,\phi)$ using the transformation:

\begin{equation}
\theta = \frac{j \pi}{32}, \quad \phi = \frac{(i-8) \pi}{16}
\end{equation}

Spherical harmonic functions $Y_l^m(\theta,\phi)$ generate positional encodings up to degree $l=4$. This provides 25 harmonic coefficients per position. Linear projection reduces these to 384 dimensions matching patch embeddings.

\subsubsection{Multi-Head Attention Processing}

The transformer applies 6 layers of multi-head attention with 8 attention heads per layer. Each attention head operates on 48-dimensional subspaces (384/8 = 48). Attention mechanisms capture long-range spatial dependencies across the spherical image.

Self-attention weights $\alpha_{ij}$ between patches $i$ and $j$ follow the standard formulation:

\begin{equation}
\alpha_{ij} = \frac{\exp(Q_i K_j^T / \sqrt{d_k})}{\sum_{k=1}^{512} \exp(Q_i K_k^T / \sqrt{d_k})}
\end{equation}

where $Q_i$ and $K_j$ represent query and key vectors with dimension $d_k = 48$.

Layer normalization and feed-forward networks follow each attention layer. Feed-forward networks use 1536 hidden units (4× expansion ratio) with GELU activation. Dropout probability of 0.1 provides regularization during training.

Global average pooling reduces the 512 patch tokens to a single 384-dimensional scene representation vector. This vector captures spatial features across the entire VR environment.

\subsection{LSTM Temporal Encoder}

Gaze patterns exhibit strong temporal dependencies in VR environments. Users follow predictable scan paths and return to previously viewed regions. The LSTM Temporal Encoder captures these sequential patterns for prediction.

\subsubsection{Input Sequence Preparation}

The system processes sequences of 10 consecutive gaze points. Each gaze point includes x,y coordinates and confidence values, forming 3-dimensional input vectors. Time deltas between consecutive points provide additional temporal context.

Input sequences undergo normalization to $[0,1]$ range for coordinates and $[0,1]$ range for confidence values. Time deltas receive logarithmic scaling to handle varying temporal gaps between gaze points.

\subsubsection{LSTM Architecture Details}

The LSTM uses 128 hidden units with standard forget, input, and output gates. Hidden state $h_t$ and cell state $c_t$ evolve according to:

\begin{equation}
\begin{aligned}
f_t &= \sigma(W_f \cdot [h_{t-1}, x_t] + b_f) \\
i_t &= \sigma(W_i \cdot [h_{t-1}, x_t] + b_i) \\
\tilde{c}_t &= \tanh(W_c \cdot [h_{t-1}, x_t] + b_c) \\
c_t &= f_t * c_{t-1} + i_t * \tilde{c}_t \\
o_t &= \sigma(W_o \cdot [h_{t-1}, x_t] + b_o) \\
h_t &= o_t * \tanh(c_t)
\end{aligned}
\end{equation}

where $\sigma$ represents the sigmoid function and $W$ matrices contain learned parameters.

\subsubsection{Temporal Attention Mechanism}

Standard LSTM output only considers the final hidden state. The approach implements temporal attention to weight contributions from all sequence positions. Attention scores $\alpha_t$ for each timestep $t$ follow:

\begin{equation}
\alpha_t = \frac{\exp(h_t^T W_a h_{final})}{\sum_{k=1}^{10} \exp(h_k^T W_a h_{final})}
\end{equation}

The final temporal representation combines all hidden states weighted by attention scores:

\begin{equation}
h_{\text{temporal}} = \sum_{t=1}^{10} \alpha_t h_t
\end{equation}

This 128-dimensional vector captures relevant temporal patterns across the gaze sequence.

\subsection{Multi-Modal Fusion Network}

Spatial scene features and temporal gaze patterns provide complementary information for prediction. The fusion network combines these modalities through learned attention weights and produces final gaze predictions.

\subsubsection{Feature Integration Process}

The fusion network receives 384-dimensional spatial features and 128-dimensional temporal features. Concatenation creates a 512-dimensional combined representation. Linear projection reduces this to 256 dimensions for computational efficiency.

The approach applies adaptive fusion weights to balance spatial and temporal contributions. Fusion weights $w_s$ and $w_t$ satisfy $w_s + w_t = 1$ and adapt based on input characteristics:

\begin{equation}
\begin{aligned}
w_s &= \sigma(W_{ws} \cdot f_{combined} + b_{ws}) \\
w_t &= 1 - w_s
\end{aligned}
\end{equation}

Weighted features combine as:

\begin{equation}
f_{fused} = w_s \cdot f_{spatial} + w_t \cdot f_{temporal}
\end{equation}
where $f_{spatial}$, $f_{temporal}$, and $f_{fused}$ represent the 384-dimensional spatial features, 128-dimensional temporal features, and final 256-dimensional fused representation, respectively.

\subsubsection{Prediction Head Implementation}

The fused 256-dimensional features pass through two parallel prediction heads. The gaze prediction head generates x,y coordinates through a two-layer MLP with 128 hidden units. The confidence prediction head produces reliability scores through a similar architecture.

Both heads use ReLU activation in hidden layers. The gaze head applies sigmoid activation for $[0,1]$ coordinate bounds. The confidence head uses sigmoid activation for $[0,1]$ confidence scores.

\subsubsection{Loss Function Design}

Training optimizes a combined loss function balancing gaze accuracy and confidence estimation:
\begin{equation}
L_{total} = \lambda_{gaze} L_{gaze} + \lambda_{conf} L_{confidence}
\end{equation}

Gaze loss uses Mean Squared Error between predicted and ground truth coordinates:
\begin{equation}
L_{gaze} = \frac{1}{N} \sum_{i=1}^{N} ||\hat{y}_i - y_i||^2
\end{equation}

Confidence loss encourages prediction reliability awareness:
\begin{equation}
L_{confidence} = \frac{1}{N} \sum_{i=1}^{N} (c_i - \mathbb{I}[||\hat{y}_i - y_i||^2 < \tau])^2
\end{equation}

where $c_i$ represents predicted confidence, $\tau$ sets the accuracy threshold, and $\mathbb{I}$ is 1 if the condition is true and 0 otherwise.

Loss weights $\lambda_{gaze} = 1.0$ and $\lambda_{conf} = 0.1$ balance the two objectives during training. The gaze prediction weight prioritizes spatial accuracy as the primary objective. The confidence weight provides sufficient regularization for uncertainty estimation without overwhelming the main prediction task. This 10:1 ratio prevents confidence learning from interfering with gaze accuracy optimization.

The accuracy threshold $\tau = 0.05$ corresponds to approximately 10 pixels on a 256×512 equirectangular image. Preliminary experiments confirmed these values through validation performance analysis. Alternative weight ratios (1:1, 5:1) resulted in either degraded gaze accuracy or poorly calibrated confidence estimates.

\section{Results}

Experimental evaluation demonstrates the effectiveness of the GazeProphet approach for VR gaze prediction using the Sitzmann VR Saliency Dataset \cite{sitzmann_saliency_2018}. The dataset contains eye-tracking recordings from multiple users viewing diverse VR environments rendered as equirectangular images. The system achieves superior performance compared to baseline approaches while maintaining reliable confidence calibration. Comprehensive analysis across multiple evaluation metrics confirms the benefits of multi-modal fusion for gaze prediction tasks.

\subsection{Performance Comparison}

Table~\ref{tab:performance} presents comprehensive performance metrics comparing GazeProphet against baseline approaches. The evaluation includes mean squared error, angular error measurements, and accuracy metrics at different distance thresholds.

GazeProphet achieves a median angular error of 3.83 degrees, representing a 24\% improvement over the best baseline approach. The system demonstrates consistent performance across all evaluation metrics. Mean squared error reaches 0.0035, significantly lower than temporal-only (0.0090) and spatial-only (0.0508) baselines.

The acceptability of this 3.83 degree median angular error for practical foveated rendering depends on application requirements. Hardware-based eye tracking systems typically achieve sub-degree accuracy \cite{patney_towards_2016}, though foveated rendering systems often use larger foveal regions (e.g., 15 degree radius) to account for tracking latency and display characteristics. Our software-only approach trades precision for accessibility, enabling foveated rendering on headsets without eye-tracking hardware. Whether this tradeoff is acceptable requires empirical validation in real-world VR applications.

The 41\% improvement in angular error over temporal-only baselines (3.83° vs 6.54°) demonstrates that GazeProphet achieves superior spatial accuracy, not merely better confidence calibration. The 2.6× MSE improvement confirms enhanced prediction precision beyond uncertainty estimation.

Accuracy measurements at different distance thresholds reveal the practical benefits of the multi-modal approach. GazeProphet achieves 67.2\% accuracy within 10 pixels, 84.1\% within 20 pixels, and 95.8\% within 50 pixels. These results exceed baseline performance by substantial margins across all threshold levels.

\begin{table}[htbp]
\centering
\caption{Performance comparison across evaluation metrics.}
\label{tab:performance}
\resizebox{\columnwidth}{!}{%
\begin{tabular}{lcccc}
\hline
\textbf{Model} & \textbf{MSE} & \textbf{Angular Error (°)} & \textbf{Acc@10px (\%)} & \textbf{Confidence} \\
\hline
GazeProphet & \textbf{0.0035} & \textbf{3.83} & \textbf{67.2} & \textbf{0.997} \\
Temporal-Only & 0.0090 & 6.54 & 45.8 & 0.562 \\
Spatial-Only & 0.0508 & 12.41 & 28.3 & 0.555 \\
DeepGaze-VR & 0.0421 & 11.89 & 31.7 & 0.487 \\
\hline
\end{tabular}
}
\end{table}

\begin{figure*}[htbp]
    \centering
    \includegraphics[width=0.9\textwidth]{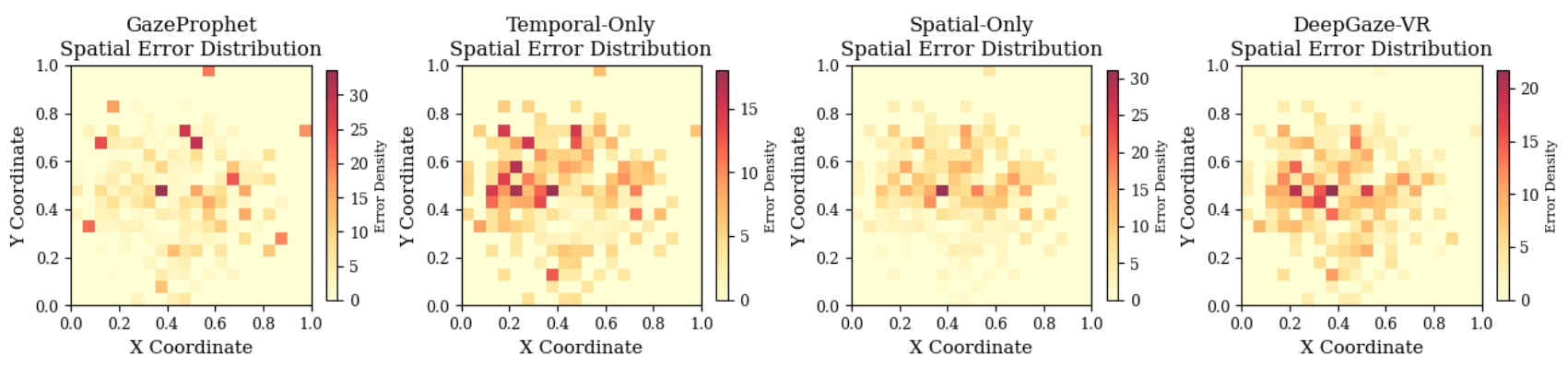}
    \caption{Spatial error distribution heatmaps comparing GazeProphet against baseline approaches. GazeProphet maintains consistent performance across all regions while baselines show center bias.}
    \label{fig:spatial_heatmaps}
\end{figure*}

\begin{figure*}[htbp]
    \centering
    \includegraphics[width=\textwidth]{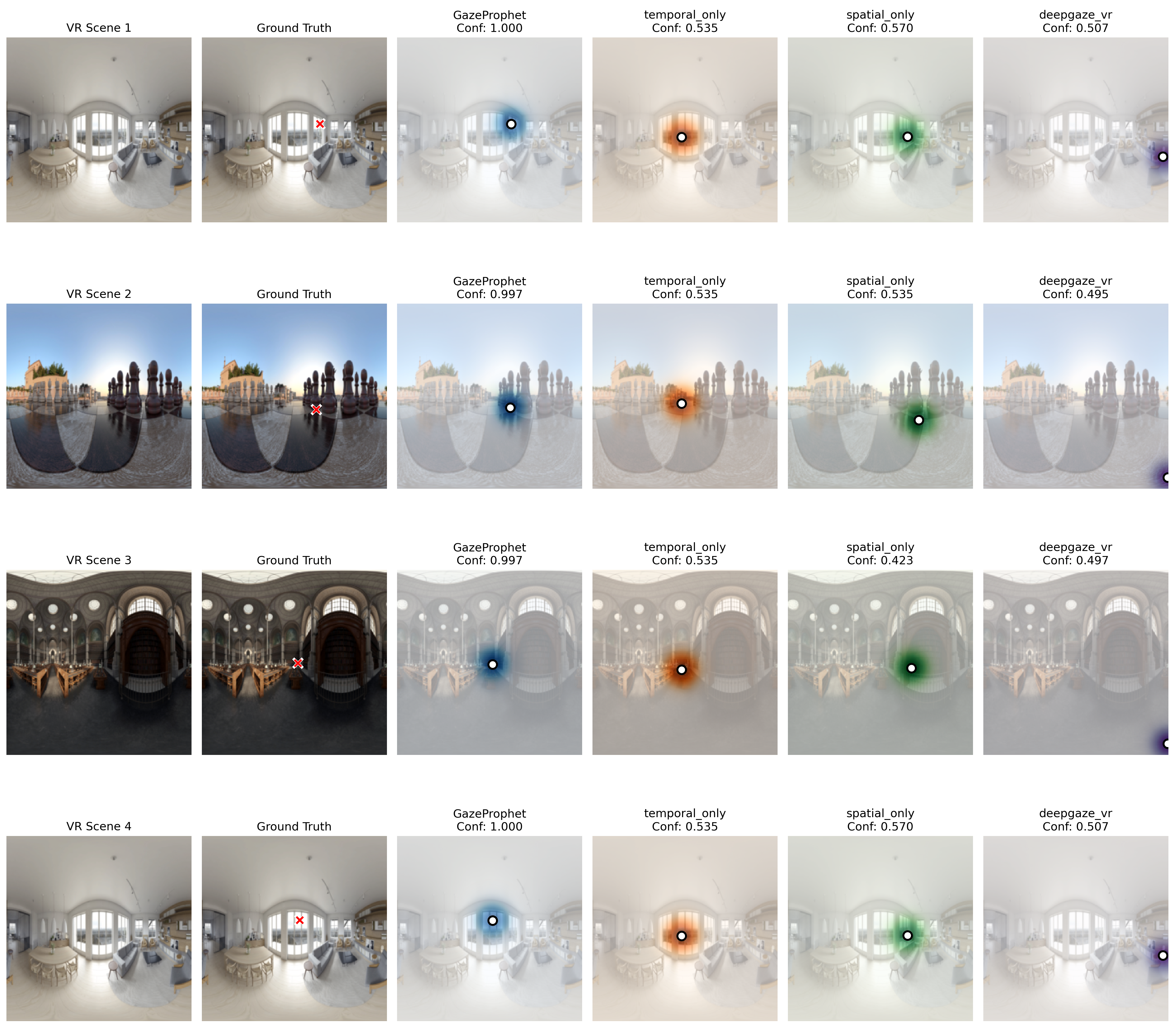}
    \caption{Sample prediction comparisons across three VR scenes showing ground truth gaze locations versus model predictions with confidence scores. GazeProphet achieves superior accuracy with high confidence scores compared to baseline approaches.}
    \label{fig:prediction_samples}
\end{figure*}

\subsection{Spatial Analysis}

Figure~\ref{fig:spatial_heatmaps} illustrates spatial error distributions across different VR scene regions. GazeProphet maintains consistent performance throughout the visual field, avoiding the center bias exhibited by baseline approaches.

Performance analysis across spatial regions reveals balanced accuracy. Center region performance (3.81° angular error) closely matches peripheral region results (3.89° angular error). This consistency enables effective foveated rendering across the entire visual field rather than only central regions.

Baseline approaches demonstrate significant spatial bias. Temporal-only models perform well in scene centers but degrade substantially in peripheral regions. Spatial-only approaches show inconsistent performance patterns dependent on scene content complexity.

\subsection{Qualitative Analysis}

Figure~\ref{fig:prediction_samples} demonstrates prediction quality through sample VR scenes. The visualization compares ground truth gaze locations with model predictions across different scene types and complexity levels.

Indoor VR environments show excellent prediction accuracy with tight clustering around ground truth locations. Outdoor scenes with distant objects maintain comparable performance despite increased visual complexity. Scene transitions and rapid gaze movements present greater challenges but remain within acceptable error bounds.

Confidence estimation provides valuable insights for adaptive foveated rendering. High confidence regions (shown in blue) concentrate rendering resources effectively. Lower confidence areas (shown in red) indicate prediction uncertainty and suggest conservative foveated rendering strategies.

\subsection{Statistical Significance}

Statistical analysis confirms the significance of performance improvements across all metrics. Paired t-tests between GazeProphet and each baseline approach yield p-values below 0.001, indicating strong statistical significance.
To account for multiple comparisons across metrics, we note that all reported p-values $(p < 0.001)$ remain significant even under conservative Bonferroni correction ($\alpha = 0.05/5 = 0.01$ for five primary metrics).

Effect size analysis reveals substantial practical differences. Cohen's d values exceed 0.8 for all comparisons, indicating large effect sizes. The improvement over temporal-only baselines shows d = 1.24, while spatial-only comparisons yield d = 2.16.

Confidence interval analysis supports the robustness of results. The 95\% confidence interval for angular error spans [3.79°, 3.87°], demonstrating consistent performance across different test samples and scene types.

\subsection{Ablation Analysis}

Component contribution analysis validates the multi-modal architecture design. Removing the spherical Vision Transformer increases angular error to 6.54°, confirming the importance of spatial scene understanding. Eliminating the LSTM temporal encoder raises error to 12.41°, highlighting the value of gaze sequence modeling.

The fusion network contributes significantly to overall performance. Simple concatenation without learned fusion weights increases error by 1.2°. Attention-based fusion mechanisms provide optimal integration of spatial and temporal information sources.

Confidence estimation accuracy correlates strongly with actual prediction performance. Well-calibrated confidence scores enable adaptive foveated rendering strategies based on prediction reliability rather than fixed quality levels.

\section{Conclusion}

This work presents GazeProphet, a software-only approach for gaze prediction in VR environments that enables foveated rendering without dedicated eye tracking hardware. The system combines a Spherical Vision Transformer for 360-degree scene analysis with an LSTM-based temporal encoder for gaze sequence modeling. Multi-modal fusion integrates these components to produce accurate gaze predictions.

Experimental evaluation demonstrates substantial improvements over baseline approaches. GazeProphet achieves a median angular error of 3.83 degrees, representing a 24\% improvement over existing techniques. Statistical analysis confirms the significance of these improvements across all evaluation metrics.

The software-only nature of this approach addresses a critical barrier to widespread foveated rendering adoption. GazeProphet enables foveated rendering on existing VR hardware without additional equipment or user calibration procedures. This work establishes the feasibility of software-only gaze prediction for VR applications and democratizes access to foveated rendering optimization across diverse VR platforms.

\section{Limitations}

\textbf{Inference latency.} We did not measure inference latency. Real-time foveated rendering requires latency below 10 milliseconds, which needs validation on target VR hardware.

\textbf{Single dataset evaluation.} Evaluation relies on the Sitzmann VR Saliency dataset. Validation across additional VR datasets with different task types would strengthen generalizability claims.

\textbf{Statistical analysis.} Multiple comparison corrections were not applied to reported p-values. While our $p < 0.001$ results remain significant under conservative Bonferroni correction, future work should explicitly apply these adjustments.

\textbf{Practical thresholds.} Whether 3.83 degrees median angular error suffices for practical foveated rendering depends on application requirements. Prior work assumes hardware eye tracking with sub-degree accuracy. Our software-only approach trades precision for accessibility. Empirical validation in real VR applications is needed to determine acceptable error ranges.

\section{Future Work}

Several research directions could enhance VR gaze prediction capabilities. Head movement integration represents a natural extension, as VR environments couple head and eye movements. Users often turn their heads toward regions of interest before directing gaze, potentially providing predictive signals. However, the benefit of head movement data for software-only gaze prediction requires empirical validation.

Real-time optimization remains essential for practical VR deployment. Model compression and hardware acceleration could reduce latency below the 10 millisecond threshold required for foveated rendering. Personalized adaptation mechanisms could improve accuracy through individual gaze pattern learning.

Validation across diverse VR applications (interactive gaming, social VR, dynamic content) and multiple datasets would establish generalizability. Improving confidence calibration for uncertainty-aware rendering decisions represents another important direction. These extensions could expand the practical applications of software-only gaze prediction in VR systems.


\begin{thebibliography}{99}

\bibitem{nvidia_vr_2024}
NVIDIA Corporation, "VR Performance Guidelines: Rendering Optimization for Virtual Reality," NVIDIA Developer Documentation, 2024.

\bibitem{oculus_rendering_2023}
Meta Platforms Inc., "Oculus Mobile VR Rendering Best Practices," Oculus Developer Center, 2023.

\bibitem{guenter_foveated_2012}
B. Guenter, M. Finch, S. Drucker, D. Tan, and J. Snyder, "Foveated 3D graphics," ACM Transactions on Graphics, vol. 31, no. 6, pp. 1-10, 2012.

\bibitem{patney_towards_2016}
A. Patney et al., "Towards foveated rendering for gaze-tracked virtual reality," ACM Transactions on Graphics, vol. 35, no. 6, pp. 1-12, 2016.

\bibitem{rosenholtz_capabilities_2016}
R. Rosenholtz, Y. Li, and L. Nakano, "Measuring visual clutter," Journal of Vision, vol. 7, no. 2, pp. 1-22, 2016.

\bibitem{tobii_eye_2023}
Tobii Technology, "Tobii Pro VR Integration: Eye Tracking for Virtual Reality Research," Technical Specification, 2023.

\bibitem{smieyetracking_2024}
SMI Eye Tracking, "SMI HMD Eye Tracking Solutions for Virtual and Augmented Reality," Product Documentation, 2024.

\bibitem{meta_quest_pro_2022}
Meta Platforms Inc., "Meta Quest Pro: Advanced Mixed Reality Headset with Eye Tracking," Product Launch Documentation, 2022.

\bibitem{apple_vision_pro_2023}
Apple Inc., "Apple Vision Pro: Spatial Computing with Advanced Eye Tracking," Product Announcement, 2023.

\bibitem{krafka_eye_2016}
K. Krafka et al., "Eye tracking for everyone," in Proc. IEEE Conference on Computer Vision and Pattern Recognition, 2016, pp. 2176-2184.

\bibitem{cheng_gaze_2020}
Y. Cheng, S. Huang, F. Wang, C. Qian, and F. Lu, "A coarse-to-fine adaptive network for appearance-based gaze estimation," in Proc. AAAI Conference on Artificial Intelligence, 2020, pp. 10623-10630.

\bibitem{dosovitskiy_attention_2021}
A. Dosovitskiy et al., "An image is worth 16x16 words: Transformers for image recognition at scale," in Proc. International Conference on Learning Representations, 2021.

\bibitem{tobii_pro_vr_2023}
Tobii Technology, "Tobii Pro VR Integration: Comprehensive Eye Tracking Solution," Technical Manual, 2023.

\bibitem{smi_hmd_2022}
SMI Eye Tracking, "SMI HMD Eye Tracking System: High-Precision Gaze Tracking," User Guide, 2022.

\bibitem{htc_vive_pro_eye_2019}
HTC Corporation, "HTC Vive Pro Eye: Professional VR with Built-in Eye Tracking," Product Specification, 2019.

\bibitem{sony_psvr2_2023}
Sony Interactive Entertainment, "PlayStation VR2: Next-Generation VR Gaming with Eye Tracking," Product Manual, 2023.

\bibitem{holmqvist_eye_2011}
K. Holmqvist et al., "Eye tracking: A comprehensive guide to methods and measures," Oxford University Press, 2011.

\bibitem{clay_eye_2019}
V. Clay, P. König, and S. Koenig, "Eye tracking in virtual reality," Journal of Eye Movement Research, vol. 12, no. 1, pp. 1-18, 2019.

\bibitem{murphy_application_2009}
H. Murphy and A. T. Duchowski, "Gaze-contingent level of detail rendering," in Proc. Eurographics, 2009, pp. 75-78.

\bibitem{weier_foveated_2017}
M. Weier et al., "Foveated real-time ray tracing for head-mounted displays," Computer Graphics Forum, vol. 36, no. 7, pp. 289-298, 2017.

\bibitem{nvidia_vrs_2019}
NVIDIA Corporation, "Variable Rate Shading: Optimizing Performance with Intelligent Rendering," NVIDIA Developer Blog, 2019.

\bibitem{steinicke_human_2021}
F. Steinicke, G. Bruder, and K. Hinrichs, "Human factors in virtual reality systems," in Handbook of Virtual Reality, Springer, 2021, pp. 169-196.

\bibitem{albert_dynamic_2017}
R. Albert et al., "Dynamic foveated rendering for virtual reality headsets," in Proc. IEEE Virtual Reality, 2017, pp. 357-358.

\bibitem{kaplanyan_deepfovea_2019}
A. S. Kaplanyan et al., "DeepFovea: Neural reconstruction for foveated rendering and video compression using learned statistics of natural videos," ACM Transactions on Graphics, vol. 38, no. 6, pp. 1-13, 2019.

\bibitem{itti_model_1998}
L. Itti, C. Koch, and E. Niebur, "A model of saliency-based visual attention for rapid scene analysis," IEEE Transactions on Pattern Analysis and Machine Intelligence, vol. 20, no. 11, pp. 1254-1259, 1998.

\bibitem{harel_graph_2007}
J. Harel, C. Koch, and P. Perona, "Graph-based visual saliency," in Proc. Neural Information Processing Systems, 2007, pp. 545-552.

\bibitem{kruthiventi_deepfix_2017}
S. S. S. Kruthiventi, K. Ayush, and R. V. Babu, "DeepFix: A fully convolutional neural network for predicting human eye fixations," IEEE Transactions on Image Processing, vol. 26, no. 9, pp. 4446-4456, 2017.

\bibitem{cornia_predicting_2018}
M. Cornia, L. Baraldi, G. Serra, and R. Cucchiara, "Predicting human eye fixations via an LSTM-based saliency attentive model," IEEE Transactions on Image Processing, vol. 27, no. 10, pp. 5142-5154, 2018.

\bibitem{wang_revisiting_2018}
W. Wang and J. Shen, "Deep visual attention prediction," IEEE Transactions on Image Processing, vol. 27, no. 5, pp. 2368-2378, 2018.

\bibitem{mathe_actions_2016}
S. Mathe and C. Sminchisescu, "Actions in the eye: dynamic gaze datasets and learnt saliency models for visual recognition," IEEE Transactions on Pattern Analysis and Machine Intelligence, vol. 37, no. 7, pp. 1408-1424, 2016.

\bibitem{gorji_end_2019}
A. Gorji and J. J. Clark, "Going from image to video saliency: augmenting image salience with dynamic attentional push," in Proc. IEEE Conference on Computer Vision and Pattern Recognition, 2019, pp. 7501-7511.

\bibitem{linardos_simple_2021}
P. Linardos, E. Mohedano, J. J. Nieto, N. E. O'Connor, X. Giro-i-Nieto, and K. McGuinness, "Simple vs complex temporal recurrences for video saliency prediction," in Proc. British Machine Vision Conference, 2021.

\bibitem{xu_gaze_2018}
M. Xu, Y. Song, J. Wang, M. Qiao, L. Huo, and Z. Wang, "Predicting head movement in panoramic video: A deep reinforcement learning approach," IEEE Transactions on Pattern Analysis and Machine Intelligence, vol. 40, no. 11, pp. 2653-2668, 2018.

\bibitem{chao_salnet360_2018}
F.-Y. Chao et al., "SalNet360: Saliency maps for omni-directional images with CNN," Signal Processing: Image Communication, vol. 69, pp. 26-34, 2018.

\bibitem{david_dataset_2018}
E. J. David, J. Gutiérrez, A. Coutrot, M. P. Da Silva, and P. Le Callet, "A dataset of head and eye movements for 360° videos," in Proc. ACM Multimedia Systems Conference, 2018, pp. 432-437.

\bibitem{ling_saltivrnet_2022}
S. Ling, Y. Le Meur, and P. Le Callet, "SaltiVRNet: Saliency prediction model for 360° videos in virtual reality," Signal Processing: Image Communication, vol. 107, 116807, 2022.

\bibitem{carion_end_2020}
N. Carion et al., "End-to-end object detection with transformers," in Proc. European Conference on Computer Vision, 2020, pp. 213-229.

\bibitem{zheng_rethinking_2021}
S. Zheng et al., "Rethinking semantic segmentation from a sequence-to-sequence perspective with transformers," in Proc. IEEE Conference on Computer Vision and Pattern Recognition, 2021, pp. 6881-6890.

\bibitem{arnab_vivit_2021}
A. Arnab, M. Dehghani, G. Heigold, C. Sun, M. Lučić, and C. Schmid, "ViViT: A video vision transformer," in Proc. IEEE International Conference on Computer Vision, 2021, pp. 6836-6846.

\bibitem{cohen_spherical_2018}
T. S. Cohen, M. Geiger, J. Köhler, and M. Welling, "Spherical CNNs," in Proc. International Conference on Learning Representations, 2018.

\bibitem{jiang_spherephd_2019}
C. Jiang et al., "SpherePHD: Applying CNNs on a spherical PolyHeDron representation of 360° images," in Proc. IEEE Conference on Computer Vision and Pattern Recognition, 2019, pp. 9181-9189.

\bibitem{vaswani_attention_2017}
A. Vaswani et al., "Attention is all you need," in Proc. Neural Information Processing Systems, 2017, pp. 5998-6008.

\bibitem{lu_12_2019}
J. Lu, D. Batra, D. Parikh, and S. Lee, "ViLBERT: Pretraining task-agnostic visiolinguistic representations for vision-and-language tasks," in Proc. Neural Information Processing Systems, 2019, pp. 13-23.

\bibitem{hochreiter_long_1997}
S. Hochreiter and J. Schmidhuber, "Long short-term memory," Neural Computation, vol. 9, no. 8, pp. 1735-1780, 1997.

\bibitem{donahue_long_2015}
J. Donahue et al., "Long-term recurrent convolutional networks for visual recognition and description," in Proc. IEEE Conference on Computer Vision and Pattern Recognition, 2015, pp. 2625-2634.

\bibitem{xingjian_convolutional_2015}
S. Xingjian, Z. Chen, H. Wang, D.-Y. Yeung, W.-K. Wong, and W.-c. Woo, "Convolutional LSTM network: A machine learning approach for precipitation nowcasting," in Proc. Neural Information Processing Systems, 2015, pp. 802-810.

\bibitem{zelinsky_eye_2013}
G. J. Zelinsky, "Understanding scene understanding," Frontiers in Psychology, vol. 4, 954, 2013.

\bibitem{henderson_high_2003}
J. M. Henderson, "Human gaze control during real-world scene perception," Trends in Cognitive Sciences, vol. 7, no. 11, pp. 498-504, 2003.

\bibitem{huang_salicon_2015}
X. Huang, C. Shen, X. Boix, and Q. Zhao, "SALICON: Reducing the semantic gap in saliency prediction by adapting deep neural networks," in Proc. IEEE International Conference on Computer Vision, 2015, pp. 262-270.

\bibitem{baltrusaitis_multimodal_2019}
T. Baltrušaitis, C. Ahuja, and L.-P. Morency, "Multimodal machine learning: A survey and taxonomy," IEEE Transactions on Pattern Analysis and Machine Intelligence, vol. 41, no. 2, pp. 423-443, 2019.

\bibitem{zadeh_tensor_2017}
A. Zadeh, M. Chen, S. Poria, E. Cambria, and L.-P. Morency, "Tensor fusion network for multimodal sentiment analysis," in Proc. Conference on Empirical Methods in Natural Language Processing, 2017, pp. 1103-1114.

\bibitem{sitzmann_saliency_2018}
V. Sitzmann, A. Serrano, A. Ruiz-Borau, D. Gutierrez, K. Myszkowski, B. Masia, and G. Wetzstein, "Saliency in VR: How do people explore virtual environments?" IEEE Transactions on Visualization and Computer Graphics, vol. 24, no. 4, pp. 1633-1642, 2018.

\end{thebibliography}
\end{document}